\documentclass{article}
\usepackage{wrapfig}
\usepackage[final]{corl_2026} 
\usepackage{amsmath, amssymb, booktabs, graphicx, svg, multirow, cleveref}
\usepackage{algorithm, algpseudocode}

\title{Your Model Already Knows: Attention-Guided Safety Filter for Vision-Language-Action Models}

\author{
  Seongbin Park, Fan Zhang, Baharan Mirzasoleiman, Shahriar Talebi, Nader Sehatbakhsh\\
  University of California Los Angeles 
  United States\\
  \texttt{parkseongbin@ucla.edu} \\
}

\providecommand{\methodname}{\textsc{Knows}}

\begin{document}
\maketitle


\begin{abstract}
    Vision-Language-Action (VLA) models have demonstrated impressive end-to-end performance across a variety of robotic manipulation tasks. However, these policies offer no guarantees against collisions with task-irrelevant objects in the scene. Existing safety filters sidestep this problem by querying a vision-language model (VLM) to identify obstacles and their locations. This, however, is too slow to run in the control loop and can only be invoked at episode initialization, leaving the filter unable to track moving obstacles. We discover that a small number of attention heads within a VLA model reliably localize the object the policy intends to approach. These heads can be exploited within a training-free safety framework that obtains the active target from the attention heads at every step, treats the remainder of the scene as obstacles, and feeds these into a Control Barrier Function (CBF) filter. Together with a lightweight real-time object tracker, this allows for collision avoidance for non-static obstacles. We evaluate our framework on \textsc{SafeLIBERO}, which we extend with moving obstacles. On the original static benchmark, our method performs comparably to an oracle that uses privileged simulator state to identify the target, emulating a VLM-based identification step run once at episode initialization. On the dynamic variant, where the oracle's init-time target assignment becomes stale, our method substantially outperforms it by 43\%, on average. Our findings suggest that the perceptual signals needed for real-time safety filtering are already present within VLA policies and can be exploited without additional training or heavy auxiliary models.
\end{abstract}



\section{Introduction}
\label{sec:intro}

Following the success of large language models (LLMs) and vision-language models (VLMs), vision-language-action models (VLAs) have emerged as a framework for end-to-end robotic control. By directly mapping vision-language input to motor commands, these models have the potential to execute previously unseen instructions and effectively generalize behaviors across a diverse range of robot embodiments, scenes, skills, and objects. Recent works, such as $\pi_0, \pi_{0.5}$~\citep{black2025pi05, black2026pi0} and OpenVLA~\citep{kim2025FineTuningVisionLanguageAction, kim2025OpenVLAOpenSource}, have made substantial progress in task execution performance; however, these models often function as unconstrained black-box policies, lacking formal safety guarantees required for deployment in real-world environments.

Before deploying such policies around fragile objects, people, and shared workspaces, we must ensure that they satisfy strict safety requirements, especially for collision avoidance. One approach is to incorporate safety constraints during training through reinforcement learning~\citep{gu2024ReviewSafe,zhang2025SafeVLASafety}. However, these methods require carefully curated safety-labeled data, substantial retraining costs, and often treat safety as a soft optimization objective rather than a hard constraint, providing limited formal guarantees~\citep{hasanzadezonuzy2021LearningSafety,wang2023EnforcingHard}.

More recently, researchers have proposed inference-time safety filters~\citep{hu2025VLSAVisionLanguageAction,brunke2025SemanticallySafe,ganai2025real,santos2025updating} that avoid retraining and instead enforce formal collision-avoidance guarantees during execution. However, existing approaches rely on expensive VLM-based scene understanding pipelines to identify safety-relevant objects. Because these perception modules are computationally heavy, they are typically executed only once at episode initialization rather than continuously throughout the rollout. This assumption is problematic in realistic robotic settings where environments are dynamic: objects may move, humans may enter the workspace, and the robot itself continuously changes its spatial relationship to surrounding obstacles. As a result, initialization-only safety filters quickly become stale and can fail to provide reliable protection under changing scene conditions.

\begin{figure}[tbp]
    \centering
    \vspace{-10pt} 
    \includegraphics[width=0.92\textwidth]{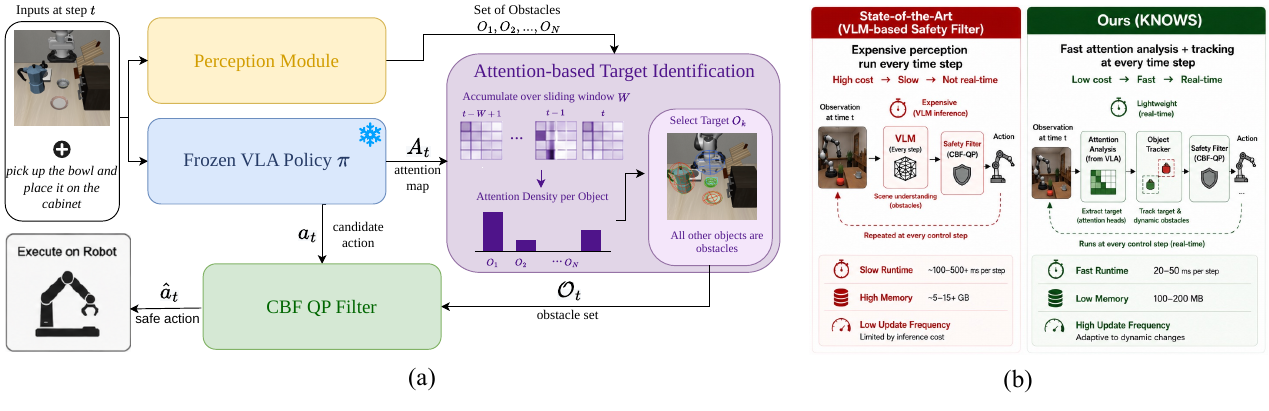}
     \vspace{-5pt}
    \caption{\textbf{(a)} To detect objects and avoid collisions using a CBF filter, our approach leverages a lightweight intra-VLA attention-based method for target identification, which eliminates the need for an expensive vision model (e.g., a VLM) for scene understanding. \textbf{(b)} Compared to state-of-the-art \citep{hu2025VLSAVisionLanguageAction,brunke2025SemanticallySafe}, our approach has lower overhead and reduces collision rates by up to 43\%.}
    \vspace{-5pt}
    \label{fig:pipeline}
\end{figure}

In this work, we observe that the information needed to identify safety-relevant objects is already encoded inside the VLA policy itself and can be extracted at negligible computational cost. Specifically, we find that a small number of attention heads in a frozen VLA consistently attend to the object the policy is currently acting toward. Rather than repeatedly querying an external VLM, we leverage these attention maps to dynamically exclude the target object from the obstacle set while treating all other objects as potential obstacles. Combined with a lightweight real-time object tracker and a control barrier function (CBF)~\citep{ames2017ControlBarrier,ames2019ControlBarrier,agrawal2017DiscreteControl} based quadratic programming (QP) solver, this yields \methodname{}\footnote{\textbf{K}nowledge-driven, \textbf{N}o-retraining, \textbf{O}nline \textbf{W}rapper for \textbf{S}afety}, a training-free safety wrapper that operates alongside a frozen VLA directly at the control rate. Unlike prior approaches, \methodname{} continuously updates its safety constraints online, enabling robust collision avoidance even in dynamic environments with moving obstacles and changing scene configurations. An overview of our approach is shown in Figure~\ref{fig:pipeline}.

We evaluate on \textsc{SafeLIBERO}~\citep{hu2025VLSAVisionLanguageAction}, a safety-augmented version of LIBERO~\citep{liu2023LIBEROBenchmarking}. To highlight the real-time capabilities of our method, we add an additional difficulty level to SafeLIBERO, where a dynamic obstacle moves adversarially during the episode.  Results show that \methodname{} reduces the collision rate by more than 43\% on average in dynamic scenarios compared to the state-of-the-art latency-heavy naive method, which can analyze the scene only once. In summary:
\begin{itemize}
  \item We identify a small set of attention heads in a frozen VLA policy that act as reliable per-step indicators of the policy's current target object, requiring no additional training or supervision. 
  \item We propose \methodname{}, a training-free safety framework that combines these target heads with a lightweight object tracker to maintain a per-step target/obstacle decomposition, which is fed into a CBF-QP filter for collision avoidance at the control rate.
  \item We extend SafeLIBERO with moving obstacles and show that \methodname{} performs on par with a privileged-state oracle on static scenes and substantially outperforms it on dynamic scenes — where init-only safety filters fail.
\end{itemize}


\section{Related Works}
\label{sec:related}
\subsection{Vision-Language-Action Models}
Following the success of VLMs~\citep{karamcheti2024PrismaticVLMs, chen2023PaLIXScaling, beyer2024PaliGemmaVersatile}, VLA models, which extend pretrained VLMs to generate low-level robot actions, have emerged as a promising generalist robot policy. The RT series~\cite{brohan2023RT1,zitkovich2023RT2} introduced action tokenization, demonstrating that scaling VL pretraining with robot datasets enables generalization across diverse manipulation tasks. Subsequent works have improved upon this end-to-end approach, including OpenVLA~\citep{kim2025OpenVLAOpenSource}, which uses autoregressive decoding, and OpenVLA-OFT~\citep{kim2025FineTuningVisionLanguageAction}, which uses parallel decoding. Another class of models utilize diffusion-based decoding, such as CogACT~\cite{li2024CogACTFoundational}, TinyVLA~\cite{wen2025TinyVLAFast} and the $\pi$ series~\citep{black2026pi0,black2025pi05}.

\subsection{Control Barrier Functions}
Control barrier functions (CBFs)~\cite{ames2017ControlBarrier,ames2019ControlBarrier} have emerged as a computationally fast and minimally invasive way of enforcing safety for non-linear systems. They have been successfully implemented on a wide range of systems, including manipulators~\cite{singletary2019online,singletary2022safety,murtaza2021real,ding2024online,morton2025safe} and discrete systems~\cite{agrawal2017DiscreteControl}. 

More recently, CBF-based safety filters have been integrated with VLA policies for robotic manipulation. \citet{hu2025VLSAVisionLanguageAction} introduces a geometric safety layer that enforces collision avoidance constraints on VLA-generated actions at runtime. \citet{brunke2025SemanticallySafe} similarly employs CBF-based shielding for semantic safety constraints to ensure safe execution without modifying the underlying policy.

\subsection{Internal Representations of Foundation Models}
Many recent works have focused on extracting spatial, semantic, and structural signals directly from the attention maps of specific heads in large foundation models~\citep{clark2019does, voita2019analyzing, michel2019sixteen}. \citet{kang2025YourLarge} discovered that frozen VLMs have a small subset of text-to-image attention heads, termed \textit{localization heads}, that implicitly capture exact object boundaries zero-shot. Transitioning this paradigm to VLA navigators, \citet{jeong2026YourVisionLanguageAction} identified specialized \textit{navigation heads}, whose spatiotemporal attention distributions provide a lightweight, training-free method for real-time path deviation and anomaly detection. In this work, we similarly identify grounding-related attention heads in manipulation VLAs and leverage their attention maps as a training-free signal for downstream safety filtering.


\section{Methodology}
\providecommand{\citeTODO}[1]{\textcolor{red}{[\textbf{cite:} #1]}}

\label{sec:method}

Our overall pipeline (as shown in Figure~\ref{fig:pipeline}a) consists of four main components: (1) A frozen VLA policy $\pi_\theta$ produces a candidate action $a_t$ at every control step (2) A perception module localizes every manipulable object in the scene: at episode initialization, we use a fine-tuned segmentation model to segment all movable objects and fit an ellipsoid over each; at runtime, the model tracks each mask to update the corresponding ellipsoid pose. (3) A target identification module reads a single attention head from $\pi_\theta$, accumulates its vision-token attention over a sliding window of recent steps, and dynamically selects the object that receives the highest accumulated attention density; the remaining objects constitute the obstacle set. (4) A CBF-QP filter takes the candidate action $a_t$ and the per-step obstacle set, and projects $a_t$ onto the safe set, yielding the final command $\hat{a}_t$ executed on the robot.

We briefly discuss the VLA policy and problem setup in \S\ref{sec:method:setup}. The details of the SAM-based ellipsoid fitting and tracking are provided in \S\ref{sec:method:perception}. In \S\ref{sec:method:attn} and \S\ref{sec:method:select}, we characterize the attention heads and how we extract a target estimate from them. \S\ref{sec:method:cbf} formulates the CBF-QP problem and the safety constraint we impose. 


\subsection{Problem Setup}
\label{sec:method:setup}

\noindent\textbf{Policy.} A pretrained VLA policy $\pi_\theta$ maps an observation $o_t$ (third-person RGB, wrist RGB, and proprioceptive state) and a language instruction $\ell$ to a chunk of $H$ actions $a_{t:t+H}$, where each action is an end-effector delta pose $(\Delta x, \Delta\theta) \in\mathbb{R}^6$ together with a gripper command. In this work, we treat $\pi_\theta$ as a black box; the model itself is neither fine-tuned nor modified.

\noindent\textbf{Scene Representation.} The scene contains a set of rigid objects $\mathcal{O}=\{1,\dots,N\}$. We represent the end-effector (EEF) and each object as a 3D ellipsoid. 
These ellipsoids $\{\mathcal{E}_j\}_{j \in \mathcal{O}}$ are fit once at the beginning of each rollout episode and tracked online, as outlined in \S\ref{sec:method:perception}. We denote the end-effector ellipsoid as $\mathcal{E}_R = (c_R, Q_R)$, where pose $(c_R, R_R)$ is read from the proprioceptive state at each timestep and the semi-axes are calibrated offline.

\noindent\textbf{Objective.} We build a safety filter that modifies the policy's nominal action $a_t$ as little as possible to a safe action $\hat{a}_t$, ensuring the end-effector ellipsoid is separated from every obstacle ellipsoid:
\begin{equation}
    \mathcal{E}_R \cap \mathcal{E}_j = \emptyset
    \qquad \forall\, j \in \mathcal{O}_t^{\mathrm{obs}}.
    \label{eq:safety_constraint}
\end{equation}
If no such $\hat{a}_t$ exists, the filter falls back to an emergency stop.

\subsection{Low-Latency Obstacle Tracking}
\label{sec:method:perception}

At episode start, an instance segmentation model~\cite{wang2025yoloe} produces a binary mask per object; we backproject the masked depth into 3D through the known camera intrinsics and extrinsics, fuse it across the available camera views into a single point cloud, and fit a minimum-volume enclosing ellipsoid (MVEE)~\cite{khachiyan1990complexity} to obtain $\mathcal{E}_i$.

Crucially, all ellipsoid shape matrices $Q_{1,2...N}$ are fixed at $t=0$. At each subsequent step, we recompute only the centroid $p_i$ from the updated segmentation mask, which avoids the cost of re-fitting an MVEE each frame. Tracks are kept stable under (i) arm occlusion, by freezing a track at its last position when the arm intervenes, and (ii) identity swaps, by re-associating tracks through an HSV color-histogram (Bhattacharyya) match. The per-step geometry update is itself a lightweight centroid recompute; the dominant per-step cost is the segmentation model's forward pass (see \S\ref{sec:results:overhead}).

\subsection{Attention-Based Target Identification}
\label{sec:method:attn}

To obtain the real-time target signals without breaking the efficiency of the policy's control rate, we extract the attention grid $A_t\in\mathbb{R}^{g\times g}$ by caching the layer inputs via lightweight hidden-state hooks during the otherwise unmodified forward pass, then manually recompute the matrix product $\mathrm{softmax}(Q_{\text{act}}K_{\text{vis}}^\top/\sqrt{d})$ exclusively for the target layer's action-query and vision-key blocks. 

After extracting the attention matrix $A_t$, each ellipsoid is projected onto the image plane and its convex hull rasterized into a mask $M_i$ with unoccluded pixel area $\alpha_{i,t}=|M_i|$. We then assign attention mass to objects proportional to image-space coverage: for every patch $(r,c)$, object $i$ receives
\begin{equation}
  \label{eq:massassign}
  m_{i,t} \;=\; \sum_{(r,c)} \bar A_t[r,c]\; c_i(r,c),
  \quad
  c_i(r,c) = \frac{|M_i \cap \text{patch}(r,c)|}{|\text{patch}(r,c)|},
\end{equation}
where $c_i(r,c)\in[0,1]$ is the fraction of the patch covered by $M_i$. Because
masks may overlap, a partially occluded but attended object still earns its share of the mass rather than losing the whole patch to whatever sits in front
of it. Because a single frame of attention is noisy, so we accumulate over a sliding window of
the last $K$ frames and score each object by an attention \emph{density}:
\begin{equation}
  \label{eq:density}
  d_i \;=\; \Big(\textstyle\sum_{K} m_{i,t}\Big)\,\Big(\textstyle\sum_{K}\alpha_{i,t}\Big)^{\beta}
\end{equation}
which is the accumulated mass divided by accumulated projected area. Normalizing by area ($\beta{=}-1$) prevents a large or nearby object from winning merely because it  occupies more patches. We confirm a target only when the top object's lead is decisive:
\begin{equation}
  \label{eq:gap}
  \tau_t = \operatorname*{arg\,max}_i d_i
  \quad\text{if}\quad d_{(1)} - d_{(2)} \ge \delta,
  \quad\text{else}\quad \tau_t = \varnothing,
\end{equation}
where $d_{(1)}\!\ge d_{(2)}$ are the two largest densities and $\delta$ is a gap threshold. When the gap is too small to trust, no object is excluded, and the filter conservatively treats the whole scene as obstacles. The values of $K, \beta$, and $\delta$ used for evaluation in~\cref{sec:results} were determined empirically; details are in the Appendix. 

\subsection{Layer/head selection}
\label{sec:method:select}
\begin{figure}[h]
    \centering
     \vspace{-5pt}
    \includegraphics[width=0.8\textwidth]{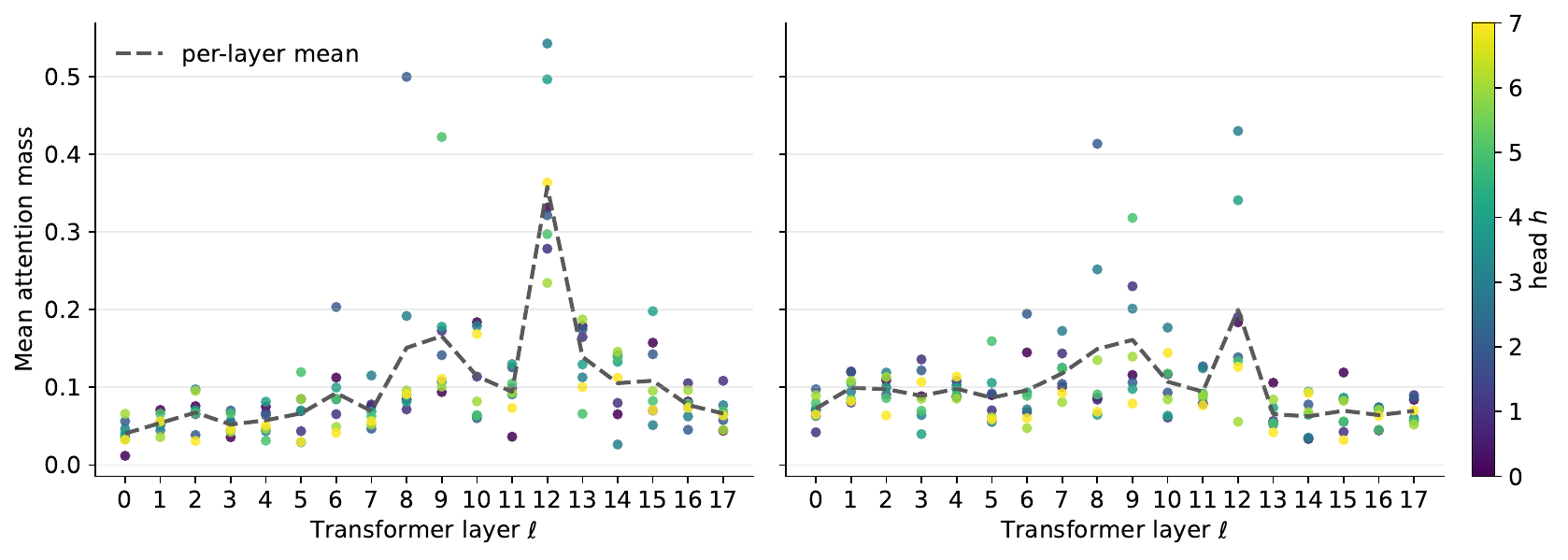}
     \vspace{-11pt}
    \caption{Per-head attention scores with policy $\pi_{0.5}$. Agent view (left) and wrist view (right).}
    \label{fig:layerhead}
     \vspace{-5pt}
\end{figure}
We run four episodes with policy $\pi_{0.5}$, one from each task in the \textsc{spatial} \textbf{I} suite (see~\cref{sec:results:expsetup} for details), while logging attention for every transformer head. To isolate the grounding signal from failure mode confounds, we deploy a ground-truth ellipsoid safety filter to guarantee collision-free trajectories during the profiling runs. For each $(\text{layer}, \text{head})$ unit, we compute the mean per-step attention mass on the phase-appropriate object (the target object during the reach phase and the destination during the placement phase). We plot the score distributions for both camera views (agent and wrist) used in our experiments in Figure~\ref{fig:layerhead}. We select the top-scoring unit (layer 12, head 3) for all suites in our main experiments.

The distributions indicate that only a small subset of heads carry strong target-identification signals, consistent with prior interpretability studies of VLMs~\citep{kang2025YourLarge}. Additionally, layer-wise averages for the agent camera peak at layer 12, followed by layer 8. The wrist camera follows the same trend, though at lower magnitudes. A possible explanation is that both camera streams are processed by the same transformer backbone and jointly trained to support action prediction, and the subset of heads for identifying task-relevant objects work regardless of viewpoint. The wrist view provides a weaker signal, likely because target objects are often outside the field of view when the end effector is distant from them.

\subsection{CBF-QP Safety Filter}
\label{sec:method:cbf}

Given the set of obstacle ellipsoids from the previous sections, we filter the policy's commanded action through a discrete-time CBF-QP. We adapt the separating-hyperplane CBF
of~\citet{wu2025CollisionfreeControl} to the case of a single controlled ellipsoid and discrete-time control. For this paper, we only consider the collision avoidance problem between the end effector ellipsoid and other obstacle ellipsoids; other components of the robot arm are not considered. See Appendix for details.

\section{Experimental Results}
\providecommand{\citeTODO}[1]{\textcolor{red}{[\textbf{cite:} #1]}}

\newcommand{\result}[1]{{--}}

\label{sec:results}

We evaluate whether attention-based target exclusion produces collision-free behavior \emph{without} sacrificing task success, and whether it does so at a cost compatible with real-time control. We organize the study around three questions: (\textbf{Q1}) can attention-based target exclusion keep collisions low without sacrificing the task success of the unfiltered policy? (\textbf{Q2}) what is the runtime overhead of reading intent from attention and filtering with the CBF? (\textbf{Q3}) what other information can be extracted from the model's attention?

\subsection{Setup}
\label{sec:results:expsetup}

\paragraph{Benchmark.}
We use \textsc{SafeLIBERO}~\citep{hu2025VLSAVisionLanguageAction}, a safety-augmented variant of
LIBERO~\citep{liu2023LIBEROBenchmarking} in which each task is populated with collision-relevant obstacles. We report on four suites, \textsc{spatial}, \textsc{object}, \textsc{goal}, and \textsc{long}, spanning tabletop, floor, and living-room arenas. Each suite is evaluated at three  safety levels: \textbf{I} (a single static obstacle close to the target object), \textbf{II} (a single static obstacle in path of movement), and \textbf{III}, our addition to the benchmark, which has dynamic obstacles that move adversarially during the episode. 

Each dynamic obstacle moves between two waypoints along a linear trajectory over 30 control steps, then stops. At our 20\,Hz control rate this corresponds to a 1.5\,s traversal. We selected this speed to be slow enough that a competent safety filter should succeed (arm motion is faster than obstacle motion), but fast enough that obstacle pose must be updated every step to avoid collisions. We run 50 episodes per task with a step budget of $300$ (\textsc{spatial}/\textsc{object}/\textsc{goal}) and $550$ (\textsc{long}) per episode. 

\noindent\textbf{Policies.}
 Our primary policy is $\pi_{0.5}$~\citep{black2025pi05}, but our safety filter is model-agnostic; the method is unchanged across policies since it reads a single attention layer/head from whatever forward pass the policy already runs. We select $\pi_{0.5}$ because it is a state-of-the-art VLA policy that exhibits strong inherent robustness to visual perturbations, which adding obstacles to a LIBERO scene inherently introduce.

\noindent\textbf{Object Tracker.} We finetune YOLOE~\citep{wang2025yoloe} to segment manipulable objects in the scene. 

\noindent\textbf{Metrics.}
 For each condition we report \emph{success rate} (SR; task completed),
\emph{collision rate} (CR; the episode contacted any non-target obstacle), and
\emph{safe-success rate} (SSR; completed \emph{and} collision-free). SSR is the
primary metric, as it jointly captures the safety/competence trade-off that
motivates target-aware filtering.

\noindent\textbf{Hardware.}
All evaluations and latency measurements were run on a single workstation with two NVIDIA RTX~PRO~6000 (Blackwell, Max-Q Workstation Edition, $96$\,GB each), an Intel Xeon w5-3425 ($12$ cores / $24$ threads), and ${\sim}\,755$\,GB of DDR5 memory, running Ubuntu~22.04. The per-step latency breakdown in
\S\ref{sec:results:overhead} is measured with the policy server on one GPU and the client (environment rendering and safety filter) on the other.

  \noindent\textbf{Baselines.} We consider three scenarios: \textbf{(i)} \textbf{No CBF.} $\pi_\theta$ executed unfiltered, with no safety layer.
  \textbf{(ii)} \textbf{Naive.} To isolate the cost of init-only obstacle estimation, we construct a naive baseline reflecting the design of VLSA \citet{hu2025VLSAVisionLanguageAction}: a single fixed obstacle ellipsoid is placed using ground-truth segmentation at $t=0$ and never updated. This is a strong stand-in for prior init-only filters and isolates the contribution of per-step updates.
  \textbf{(iii)} \textbf{\methodname{} (ours).} The full target-aware filter: every tracked object is a candidate obstacle, the
    policy's attention identifies the intended target at \emph{every} step, the target is excluded from the CBF obstacle set, and all objects are re-localized per step. 

\begin{table*}[t]
  \centering
  \setlength{\tabcolsep}{4pt}
   \caption{Main results on \textsc{SafeLIBERO} (\%). SR = success, CR = collision,
  SSR = safe-success (higher SR/SSR better, lower CR better). Best value in each metric column (per suite/level) in \textbf{bold}.} 
  \resizebox{.6\textwidth}{!}{%
  \begin{tabular}{llccccccccc}
    \toprule
    & & \multicolumn{3}{c}{Level I} & \multicolumn{3}{c}{Level II} & \multicolumn{3}{c}{Level III} \\
    \cmidrule(lr){3-5}\cmidrule(lr){6-8}\cmidrule(lr){9-11}
    Suite & Method & SR$^{\uparrow}$ & CR$^{\downarrow}$ & SSR$^{\uparrow}$ & SR$^{\uparrow}$ & CR$^{\downarrow}$ & SSR$^{\uparrow}$ & SR$^{\uparrow}$ & CR$^{\downarrow}$ & SSR$^{\uparrow}$ \\
    \midrule
    \multirow{3}{*}{\textsc{spatial}}
      & No CBF        & 67.5 & 86.5 & 13.5 & 49.0 & 88.0 & 11.0 & \textbf{86.5} & 84.5 & 14.5 \\
      & Naive         & \textbf{69.0} & \textbf{26.0} & \textbf{61.0} & 80.5 & 34.5 & 63.5 & 79.5 & 62.0 & 34.0 \\
      & \methodname{} & 63.0 & 32.5 & 53.0 & \textbf{87.5} & \textbf{27.0} & \textbf{70.0} & 66.5 & \textbf{29.0} & \textbf{54.5} \\
    \midrule
    \multirow{3}{*}{\textsc{object}}
      & No CBF        & 43.0 & 83.0 & 14.0 & 69.0 & 72.0 & 25.5 & 48.5 & 48.5 & 44.0 \\
      & Naive         & \textbf{73.0} & \textbf{4.0} & \textbf{73.0} & \textbf{87.5} & \textbf{11.0} & \textbf{80.0} & 51.0 & 50.0 & 40.0 \\
      & \methodname{} & 71.0 &  6.5 & 70.5 & 79.5 & 17.5 & 69.0 & \textbf{78.5} & \textbf{14.0} & \textbf{70.5} \\
    \midrule
    \multirow{3}{*}{\textsc{goal}}
      & No CBF        & 51.0 & 95.0 &  5.0 & 66.0 & 64.5 & 29.0 & 79.0 & 90.0 &  9.0 \\
      & Naive         & \textbf{89.5} & \textbf{9.0} & \textbf{82.0} & 55.0 & \textbf{36.0} & 36.5 & 79.0 & 90.5 &  9.5 \\
      & \methodname{} & 87.0 & 10.0 & 81.0 & \textbf{82.5} & 41.0 & \textbf{52.0} & \textbf{86.0} & \textbf{30.5} & \textbf{63.5} \\
    \midrule
    \multirow{3}{*}{\textsc{long}}
      & No CBF        & 58.5 & 86.0 & 13.5 & 47.0 & 83.5 & 14.5 & \textbf{63.5} & 82.0 & 18.0 \\
      & Naive         & \textbf{60.5} & \textbf{32.0} & \textbf{36.5} & \textbf{48.0} & \textbf{14.0} & \textbf{42.5} & 44.5 & 80.5 & 18.5 \\
      & \methodname{} & 49.0 & 45.5 & 35.0 & 35.0 & 39.5 & 24.0 & 45.5 & \textbf{34.0} & \textbf{34.5} \\
    \bottomrule
    \label{tab:main}
  \end{tabular}%
  }
 
\end{table*}

\subsection{Main Results}
\label{sec:results:main}

~\Cref{tab:main} reports SR / CR / SSR across suites and safety levels. In static obstacle scenarios (Level~I and~II) \methodname{} sharply reduces collisions relative to the unfiltered policy (No CBF) while keeping success comparable to the Naive version. In Level~III, obstacle pre-episode no longer suffices, and \methodname{} attains a higher SSR than the Naive version.
\subsection{Real-Time Overhead}
\label{sec:results:overhead}

\begin{wraptable}{r}{0.46\textwidth}
\vspace{-1.8em}
\centering
\caption{Per-step latency breakdown (mean over $200$ control steps). The attention is computed during the existing forward pass; action chunking ($H{=}8$) amortizes the policy forward to $\sim\!30$\,ms/step.}
\label{tab:overhead}
\small
\setlength{\tabcolsep}{4pt}
\begin{tabular}{lc}
\toprule
Component & Latency (ms) \\
\midrule
VLA Policy inference (one chunk) & $243$ \\
\quad attention extraction & $0.8$ \\
\midrule
YOLOe-11m-seg & $19.3$ \\
Depth + centroid & $9.1$ \\
Target identification & $9.4$ \\
Safety QP (OSQP) & $11.4$ \\
\midrule
\textbf{Wrapper overhead} & $\mathbf{49.3}$ \\
Control rate & $20$\,Hz \\
\bottomrule
\end{tabular}
\vspace{-1em}
\end{wraptable}

Target-aware safety adds little on top of the policy at runtime.
Table~\ref{tab:overhead} decomposes the per-step wall-clock cost, measured against
a $\pi_{0.5}$ server over $200$ control steps with the agentview rendered at
$640{\times}640$. Reading the target from attention is effectively free---$0.8$\,ms
per query, since it reuses the existing forward pass rather than materializing the
full attention map---and the safety QP, with one constraint per non-target
obstacle, solves in $11$\,ms. The per-step budget is dominated by the off-the-shelf
segmentation detector ($19$\,ms) and the masked-depth backprojection ($9$\,ms);
because the MVEE shape is fit only at initialization, per-step tracking is just a
centroid update. The full wrapper totals $49$\,ms---within the $50$\,ms budget of
LIBERO's $20$\,Hz control rate---and action chunking ($H{=}8$) amortizes the policy forward to $\sim\!30$\,ms/step, so the system holds control rate. Perception is the only term near budget and is readily reduced with a lighter detector, lower input resolution, or a compute optimization.

\subsection{Attention Focus as a Confidence Signal}
\label{sec:results:confidence}

\begin{figure}[htbp]
    \centering
    \vspace{-10pt} 
    \includegraphics[width=\textwidth]{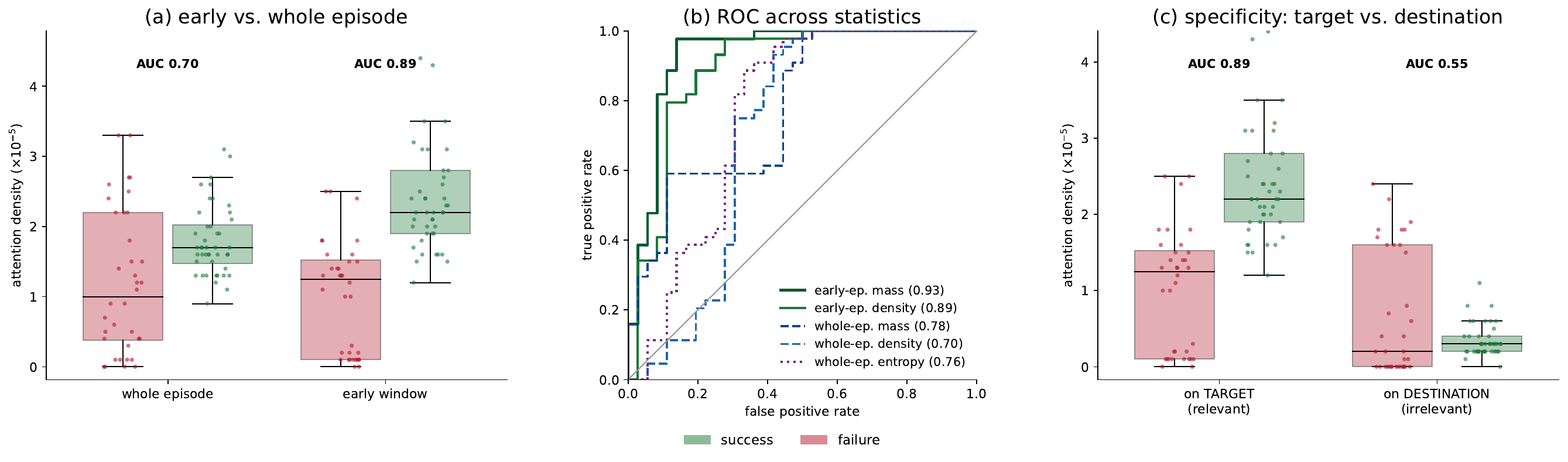}
     \vspace{-15pt}
    \caption{\textbf{Attention separates successful from failed episodes.} At the evaluation head (agent camera, layer~12, head~3) over $80$ \textsc{long} episodes ($44$ success / $36$ failure) in the analysis condition, where attention is recorded but not used for control.
  \textbf{(a)~Whole episode vs.\ early window.} Density on the phase-relevant target, restricted to the early phase, sharpens the separation from AUC~$0.70$ to $0.89$
  \textbf{(b)~ROC across statistics.} Five successful metrics overlaid: early target mass (AUC~$0.93$) and density ($0.89$); whole-episode mass ($0.78$) and density ($0.70$); and attention entropy ($0.76$). All sit well above chance.
  \textbf{(c)~Specificity.} In the early window, density on the target predicts success (AUC~$0.89$) while density on the currently-irrelevant destination is at chance (AUC~$0.55$)}
    \label{fig:confidence}
\end{figure}

We find that another useful property of the attention signal is that it is an inference-time correlate of task outcome---readable from the same forward pass that produces the action, at no extra supervision. \Cref{fig:confidence} outlines the results of analyzing the properties of the identified evaluation head (layer 12, head 3) across 80 \textsc{long} \textbf{I} episodes comprising 44 successful and 36 failed trials. In these trials, attention signatures are recorded passively without being utilized for active control loop filtering; the Naive safety filter (see~\cref{sec:results:expsetup}) was used instead.

If the signal is aggregated across the entire episode (\cref{fig:confidence}a, left), the metric yields an AUC of 0.70. However, restricting our observation to before the first object is picked up or the first third of the episode, whichever is earlier (\cref{fig:confidence}a, right), substantially sharpens the classification boundary, boosting the predictive power to an AUC of 0.89. Additionally, we evaluate five distinct internal attention statistics to determine the most robust marker for downstream safety filtering. As illustrated by the Receiver Operating Characteristic (ROC) curve in~\cref{fig:confidence}b, all five statistics perform substantially above chance (AUC=0.50).

Finally, we check whether this signal reflects semantic target alignment or just generic attention concentration. \Cref{fig:confidence}c compares early-window density on the task-relevant target against the (currently task-irrelevant) destination: the former remains strongly predictive of success (AUC=0.89) while the latter collapses to chance (AUC=0.55). The evaluation head is therefore semantically specific to the commanded object, not a generic saliency cue.

These results suggest a possible positive reinforcement cycle: since a more robust and capable base policy yields more informative attention maps, improving the policy backbone could translates into a more robust safety filter without requiring additional training.


\section{Limitations}
Our approach has several limitations that cause collisions despite the safety filter. We discuss them in this section and point to future work.

\noindent\textbf{End-effector-only safety.} The CBF protects a single ellipsoid approximating the end-effector; the rest of the arm is unmodeled. Collisions involving links upstream of the wrist are therefore not accounted for, leading to occasional upper-arm and elbow collisions. Extending the filter to cover the whole kinematic chain is the natural next step; however, our safety filter assumes control of only the end effector position and pose, leaving the rest of the joints for a downstream controller. Faithfully protecting the whole arm would require either modeling that controller's behavior or moving the safety layer to a lower level.

\noindent\textbf{Perception error.} Our obstacle geometry is only as good as the perception module that produces it. Point-cloud noise can yield ellipsoids that over- or under-approximate an object, and the tracker can mis-associate under heavy occlusion, despite our counter-measures. These errors translate directly into either overly conservative behavior or collisions.

\noindent\textbf{Reduced-order safety.} The CBF is defined over the EEF pose, since EEF deltas are the only control authority the VLA exposes. The joint trajectories that realize these
commands are produced by a downstream OSC we treat as a black box. In
the language of reduced-order safety-critical
control~\citep{cohen2025SafetyCriticalController, molnar2023SafetyCriticalControl},
the EEF is the ROM and the full joint-space system is the
full-order model; safety on the ROM transfers to the full system only
modulo the OSC's tracking error, which we validate empirically in
Section~\ref{sec:results}.


\section{Conclusion}
\label{sec:conclusion}
In this paper, we presented a minimally invasive safety filter for VLA policies utilizing the policy's own internal attention to decide which objects in the scene are obstacles. Our key finding is that even a single attention head in a frozen VLA reliably localizes the object the policy is acting towards, providing a per-step target estimate at negligible cost and without finetuning. We then treat the remainder of the tracked scene as obstacles, then utilize a CBF-QP filter to prevent collisions at a rate fast enough to run alongside the policy at control rate (20 Hz). When all objects in the scene are static, our method performs comparably to prior methods that generate and fix the obstacles prior to runtime, but outperforms then in situations where obstacles move during the episode. 

Because our approach relies entirely on the structural routing properties inherent to multi-head self-attention, this method can readily generalize to other transformer-based policies without architectural modification, though its ultimate efficacy remains tethered to the underlying model's quality; We hypothesize that a weaker backbone may produce diffuse or erratic attention maps under noise, while a robust policy like $\pi_{0.5}$ gitves useful information for filtering. 

We further observed that the same attention density our filter uses as a target signal is an inference-time correlate of eventual task success, suggesting that the policy's internal attention carries usable information well beyond the safety setting we study here. More broadly, our results indicate that the perceptual grounding needed to make a VLA safe is, to a large extent, already latent in the policy.


\clearpage
\acknowledgments{If a paper is accepted, the final camera-ready version will (and probably should) include acknowledgments. All acknowledgments go at the end of the paper, including thanks to reviewers who gave useful comments, to colleagues who contributed to the ideas, and to funding agencies and corporate sponsors that provided financial support.}


\bibliography{references}  
\clearpage
\section{Appendix}
\label{sec:appendix}
\subsection{CBF-QP Safety Filter}
\paragraph{Separating-hyperplane CBF} For two general ellipsoids $\mathcal{E}_R$ and $\mathcal{E}_O$ in $\mathbb{R}^3$, \citet{wu2025CollisionfreeControl} characterize the existence of a
separating hyperplane $\{y \mid n^\top y = \gamma\}$ through two functions,
\begin{equation}
    h_R(n, \gamma) = n^\top c_R - \gamma - \sqrt{n^\top Q_R\, n},
    \qquad
    h_O(n, \gamma) = -n^\top c_O + \gamma - \sqrt{n^\top Q_O\, n},
\end{equation}
where the square-root terms are the support functions of the two ellipsoids along $n$. The conditions $h_R \geq 0$ and $h_O \geq 0$ jointly certify that $\mathcal{E}_R$ and $\mathcal{E}_O$ lie on opposite sides of the hyperplane, and hence that $\mathcal{E}_R \cap \mathcal{E}_O = \emptyset$. The hyperplane parameters $(n, \gamma)$ are treated as virtual states, with $\|n\| = 1$ enforced throughout.

Because the obstacle ellipsoid $\mathcal{E}_O$ is uncontrolled in our setting, we eliminate the hyperplane offset $\gamma$ by summing the two functions, yielding a single combined function
\begin{equation}
    h(n) = n^\top(c_R - c_O)
           - \sqrt{n^\top Q_R\, n}
           - \sqrt{n^\top Q_O\, n}.
    \label{eq:cbf}
\end{equation}
We retain the hyperplane normal $n$ as a per-obstacle virtual state,
initialized along the center-to-center direction $c_R - c_O$ and updated incrementally within the QP below. We note that enforcing $h \geq 0$ is a relaxation of the joint condition $h_R \geq 0 \wedge h_O \geq 0$ used in~\citep{wu2025CollisionfreeControl} to certify collision-freeness; the two coincide when the hyperplane is well-positioned between the two ellipsoids, which the center-to-center initialization and the bounded per-step normal update below encourage in practice. We therefore treat \cref{eq:cbf} as a practical safety margin rather than a formal collision-free certificate, and validate its effectiveness empirically (\cref{sec:results}).

\paragraph{Discrete-time CBF constraint} At each control step we enforce the discrete-time CBF condition~\citep{agrawal2017DiscreteControl}
\begin{equation}
    \Delta h_j \;\geq\; -\gamma_h\, h_j,
    \qquad \gamma_h \in (0, 1],
    \label{eq:dt_cbf}
\end{equation}
for every obstacle $j \in \mathcal{O}_t^{\mathrm{obs}}$, where $h_j$ is
\cref{eq:cbf} evaluated for the pair $(\mathcal{E}_R, \mathcal{E}_j)$ with
the current virtual normal $n^{(j)}$. This condition keeps $h_j$ from
decreasing faster than the rate $\gamma_h$, i.e.\
$h_j(t{+}1) \geq (1 - \gamma_h)\, h_j(t)$. Linearizing $h_j$ about the current
state gives the affine constraint
\begin{equation}
    \nabla_{c_R} h_j \cdot \delta c_R
    \;+\; \nabla_{R_R} h_j \cdot \delta\theta
    \;+\; \nabla_{n^{(j)}} h_j \cdot \delta n^{(j)}
    \;\geq\; -\gamma_h\, h_j,
    \label{eq:cbf_linearized}
\end{equation}
where $\delta c_R \in \mathbb{R}^3$ and $\delta\theta \in \mathbb{R}^3$ are
the translational and rotational increments to the end-effector pose, and
$\delta n^{(j)} \in \mathbb{R}^3$ is the increment to the virtual hyperplane
normal. The gradients admit closed forms:
\begin{align}
    \nabla_{c_R} h_j     &= n^{(j)}, \\[2pt]
    \nabla_{R_R} h_j     &= \frac{n^{(j)} \times Q_R\, n^{(j)}}
                                 {\sqrt{{n^{(j)}}^{\!\top} Q_R\, n^{(j)}}}, \\[2pt]
    \nabla_{n^{(j)}} h_j &= (c_R - c_j)
        - \frac{Q_R\, n^{(j)}}{\sqrt{{n^{(j)}}^{\!\top} Q_R\, n^{(j)}}}
        - \frac{Q_j\, n^{(j)}}{\sqrt{{n^{(j)}}^{\!\top} Q_j\, n^{(j)}}}.
    \label{eq:cbf_grads}
\end{align}
The rotational gradient arises from the dependence of $Q_R$ on the
end-effector orientation and corresponds to the $d = 3$ form
of~\citet[eq.~26]{wu2025CollisionfreeControl}.

\paragraph{QP}Let $(\delta c_R^{\mathrm{nom}}, \delta\theta^{\mathrm{nom}})$ denote the
policy's nominal action scaled to physical units. The safety filter solves
\begin{equation}
\begin{aligned}
    \min_{\delta c_R,\, \delta\theta,\, \{\delta n^{(j)}\}}
        \;\;& \|\delta c_R - \delta c_R^{\mathrm{nom}}\|^2
            + W\,\|\delta\theta - \delta\theta^{\mathrm{nom}}\|^2 \\
    \text{s.t.}\;\;
        & \nabla_{c_R} h_j \cdot \delta c_R
          + \nabla_{R_R} h_j \cdot \delta\theta
          + \nabla_{n^{(j)}} h_j \cdot \delta n^{(j)}
          \geq -\gamma_h\, h_j, & \forall j \in \mathcal{O}_t^{\mathrm{obs}}, \\
        & \|\delta n^{(j)}\|_\infty \leq \epsilon, & \forall j \in \mathcal{O}_t^{\mathrm{obs}},
\end{aligned}
\label{eq:qp}
\end{equation}
where $W$ trades off rotational against translational tracking and $\epsilon$ bounds the per-step change in each virtual normal, keeping the hyperplane estimates smooth across steps. \cref{eq:qp} is a convex QP, which we solve with OSQP~\citep{stellato2018OSQPOperator}. After solving, we renormalize each virtual normal, $n^{(j)} \leftarrow (n^{(j)} + \delta n^{(j)}) / \|n^{(j)} + \delta n^{(j)}\|$, apply $(\delta c_R, \delta\theta)$ to the policy's commanded action (the gripper command passes through unmodified), and execute the result. If the QP is infeasible, we fall back to an emergency stop with zero translational and rotational deltas.

\paragraph{Reduced-order safety}The CBF in~\cref{eq:cbf} is defined over the end-effector pose $(c_R, R_R)$, which we treat as a kinematic state directly commanded by the filtered action $(\delta c_R, \delta\theta)$; the underlying robot dynamics are abstracted by a downstream operational-space controller (OSC) that tracks these commands. This layering is a standard reduced-order model (ROM) approach to safety-critical control~\citep{cohen2025SafetyCriticalController, molnar2023SafetyCriticalControl}:
the CBF-QP enforces the safety condition on the ROM, and the guarantee transfers to the full system modulo the OSC's tracking error. We do not characterize this error analytically; instead, we validate end-to-end safety through the collision-rate measurements in~\cref{sec:results}.

\subsection{Extracting attention under fused kernels}
During each policy query we obtain an attention grid $A_t\in\mathbb{R}^{g\times g}$ over the $g^2$ vision tokens of the third-person image, from a \textit{single} transformer layer $\ell$ and head $h$. This grid measures how strongly the action tokens the policy uses to decode $a_{t:t+H}$ attend to each image patch, and it is obtained without an extra model evaluation, a backward pass, or any retraining, allowing for real-time control.

However, obtaining this grid is not straightforward, as modern VLAs run attention with fused kernels (via FlashAttention~\citep{dao2024FlashAttention2Faster, dao2022FlashAttentionFast}) that evaluate $\mathrm{softmax}(QK^\top/\sqrt{d})V$ without ever materializing the
$T\times T$ attention matrix in memory. Disabling FlashAttention so that the transformer returns its attention maps incurs too big of a cost for real-time applications. We instead leave the fused forward pass untouched and recompute only the single map we need:
\begin{enumerate}
  \item We attach lightweight forward hooks to each attention module that cache its input hidden states: the vision/language tokens in the VLM prefix stack and the action tokens in the action-expert suffix stack. Caching layer inputs is negligible relative to the forward pass and leaves the fused kernel unchanged.
  \item Once the policy has returned its actions, for the chosen layer $\ell$ we re-project the queries from the cached action tokens and the keys from the cached vision tokens of the selected camera, re-apply rotary position embeddings at their absolute sequence positions, expand the key heads to match the query heads (grouped-query attention), and evaluate $\mathrm{softmax}(Q_{\text{act}}K_{\text{vis}}^\top/\sqrt{d})$ manually.
\end{enumerate}
This reconstructs attention for exactly one layer and only the
action-query\,$\times$\,vision-key block ($H\times g^2$ entries), instead of the full $T\times T$ map across all $L$ layers, so the added work is one small matrix multiply and softmax. This is negligible in comparison to the policy's full inference pass. 

\end{document}